\documentclass[conference]{IEEEtran}
\IEEEoverridecommandlockouts
\usepackage{cite}
\usepackage{amsmath,amssymb,amsfonts}
\usepackage{graphicx}
\usepackage{textcomp}
\usepackage[table]{xcolor}
\usepackage{csquotes}
\usepackage{enumitem}
\usepackage[ruled,vlined]{algorithm2e}
\usepackage{algpseudocode}

\include{pythonlisting}

\def\BibTeX{{\rm B\kern-.05em{\sc i\kern-.025em b}\kern-.08em
    T\kern-.1667em\lower.7ex\hbox{E}\kern-.125emX}}
    
\newlength\myindent
 \DeclareMathOperator*{\argmax}{argmax}

\newcommand{\rev}[2]{#2}

\usepackage{url}

\usepackage{breakurl}
\usepackage[breaklinks]{hyperref}

\begin{document}

\title{Generating and Adapting to Diverse Ad-Hoc Partners in Hanabi\\
}

\author{\IEEEauthorblockN{Rodrigo Canaan}
\IEEEauthorblockA{
\textit{ Cal Poly / NYU}\\
San Luis Obispo, USA \\
rcanaan@calpoly.edu}
\and
\IEEEauthorblockN{Xianbo Gao}
\IEEEauthorblockA{
\textit{NYU}\\
New York, USA \\
xg656@nyu.edu}
\and
\IEEEauthorblockN{Julian Togelius}
\IEEEauthorblockA{
\textit{NYU}\\
New York, USA \\
julian.togelius@nyu.edu}
\and
\IEEEauthorblockN{Andy Nealen}
\IEEEauthorblockA{
\textit{USC}\\
Los Angeles, USA \\
nealen@usc.edu}
\and
\IEEEauthorblockN{Stefan Menzel}
\IEEEauthorblockA{
\textit{HRI Europe}\\
Offenbach, Germany \\
stefan.menzel@honda-ri.de}
}

\maketitle

\begin{abstract}
Hanabi is a cooperative game that brings the problem of modeling other players to the forefront. In this game, coordinated groups of players can leverage pre-established conventions to great effect. In this paper, we focus on ad-hoc settings with no previous coordination between partners. We introduce a ``Bayesian Meta-Agent'' that maintains a belief distribution over hypotheses of partner policies. The policies that serve as initial hypotheses are generated using MAP-Elites, to ensure behavioral diversity. We evaluate an ``Adaptive'' version of the agent, which selects a response policy based on the updated belief distribution and a ``Generalist'' version, which selects a response based on the uniform prior. In short episodes of 10 games with a consistent partner, the ``Adaptive'' version outperforms the ``Generalist'' when the training and evaluation populations are the same. This presents a first step towards an agent that can model its partner and adapt within a time frame that is compatible with human interaction.
\end{abstract}

\section{Introduction}

Many of the most visible successes of game AI research have been systems (agents) for playing competitive, zero-sum games such as \textit{Deep Blue}~\cite{campbell2002deep} in Chess, \textit{Alpha Go}~\cite{silver2016mastering} in Go and \textit{Alpha Star}~\cite{vinyals2019grandmaster} in Starcraft. Designers of such systems often strive to make them \textit{robust} in the sense that it should be hard even for motivated, expert human players to find strategies that beat the one used by the system. In other words, the system's strategy is meant to approximate a Nash Equilibrium.

However, these systems are typically not \textit{adaptive}, in the sense that they are not designed to modify their policies, without designer intervention, based on a short number of games played with humans after deployment. In competitive games, adaptivity may or may not be desirable, as an adaptive agent deviating from a Nash Equilibrium may be more exploitable.

Cooperative games, on the other hand, are a domain where systems that can adapt on a time scale compatible with human play would be attractive, for example, as a built-in AI partner for a cooperative strategy game. As the player develops their understanding of the game, an agent that is able to model and adapt to the player's current strategy might not only achieve better scores but also enhance the player's enjoyment or even contribute to the player's learning process, when compared with a one-size-fits-all agent with a stationary policy.

In this paper, we provide an example of what such a system might look like. We describe a ``meta-agent'' for the cooperative card game Hanabi that is trained with a population of behaviorally diverse rule-based agents, generated using MAP-Elites. During training, the meta-agent collects information meant to help it identify its training partners, conditioned on the game history and the meta-agent's own policy.

We evaluate the meta-agent in episodes played with \rev{ad-hoc partners}{diverse ad-hoc partners}. The meta-agent uses Bayesian inference to maintain a belief distribution that assigns probabilities to the hypotheses that the current anonymous partner uses the same strategy as a given training partner. The meta-agent then selects actions according to a strategy that maximizes the expected rewards weighted by the belief distribution, which is tracked over independent episodes consisting of 10 games each.

This paper is an extension of our previous paper~\cite{canaan2019diverse} presented at the  2019 IEEE Conference on Games (CoG). The process of generating rule-based agents with MAP-Elites is based on that paper. Our novel contributions are the meta-agent itself, the introduction of a new behavioral metric (IPP) and a more detailed discussion about the relevance of techniques that account for behavioral diversity in the context of ad-hoc cooperative game AI benchmarks. 

 \rev{We show that the meta-agent is able to improve its performance over a ``Generalist'' baseline across independent series of 10 games, a number of games compatible with human interaction.}{The meta-agent is, as far as we know, the first agent for collaborative play in Hanabi that adapts by switching between policies and, to the best of our knowledge, the first that makes decisions based on information collected across multiple matches. This poses problems for evaluating the agent, as there is no established baseline to beat. In this paper, we compare the meta-agent's performance both with a ''Generalist'' (non-adaptive) and "Random-Response'' baseline. The Adaptive and Generalist versions handily outperform the Random-Response baseline, with the Adaptive version achieving slightly higher scores within its training distribution, but slightly lower when evaluated outside its training distribution.}

\section{Hanabi: the game}\label{sec:hanabi}

\subsection{Rules}

Hanabi (Bauza, 2010) is a cooperative card game that won the prestigious \textit{Spiel des Jahres} award in 2013~\cite{spielbgg}. It is played by groups of 2-5 players with a deck of 50 cards, where each card has one of five colors (\textbf{B}, \textbf{R}, \textbf{Y}, \textbf{W} and \textbf{G}) and a rank (a numeric value from \textbf{1} to \textbf{5}). The cards represent colored fireworks and the goal is to build one stack of each color by playing cards in ascending rank order. The twist of the game is that players play with their hands facing outwards, so that every player can see the rank and color of cards in their partners' hands, but not of those in their own hand.

Players alternate turns by choosing one of three actions: 
\begin{enumerate}
    \item \textbf{Play a card.} The team then scores a point if the card was played correctly (example, if the card is \textbf{B2} where \textbf{B1} has already been played), but loses a life from a shared pool if the card was incorrect (e.g. playing \textbf{B2} if no Blue cards have yet been played or if a Blue card with equal or higher rank has been played previously). Note that the player does not necessarily know the rank or color of a card before committing to playing it.
    \item \textbf{Give a hint.} This consumes a ``hint'' token from a shared pool and allows the player to pick either a color or rank, then identify all cards with that rank or color in a chosen teammate's hand (e.g. ``your first, second and fourth cards are all Red''). Note that the example also implies the other cards are not Red. This is the only method of communication allowed in the game.
    \item \textbf{Discard a card.} This replenishes a hint token for the team. Note that the number of duplicates of each card in the deck is limited, so taking this action may lead to a game state where it is not possible to fully build all piles (e.g. discarding all \textbf{3}s from a color would make the \textbf{4} and \textbf{5} of that color impossible to play in the future).
\end{enumerate}

Whenever a card is played or discarded, players must draw back to their hand limit. If the deck is exhausted, every player gets one last turn, after which the game ends. In this case, the players score from 0 to 24 points based on the number of cards successfully played. If the five stacks are completed with cards ranked 1 to 5 of each color, the team wins and scores 25 points.

The game also ends with defeat if the group loses three lives. In this paper, we use the ``lenient'' variant of the rules used in~\cite{walton2017evaluating} where the end score in this case is also equal to the number of cards successfully played. There also exists a ``strict'' variant, used in papers such as~\cite{bard2020hanabi} and most editions of the game, where this results in a score of zero. The complete rules of the game can be found at~\cite{hanabirules}. 

\subsection{Strategy and Conventions}\label{sub:conventions}

All information in the game is observable by all players except the content (card colors and ranks) of each player's own hand. The central theme of the game is that there are generally not enough hints to reveal all relevant information to every player, even accounting for discards. Therefore, players typically rely on assumptions about other players to infer additional meaning from each of their actions. 

For example, a common assumption is that hints are usually meant to be actionable: they should enable the receiver to successfully play a card, discard a card that is no longer useful or hold on to a card that might have otherwise been discarded.

Under this assumption, if Alice gives Bob a hint like ``your third card is a \textbf{2}'' and there is a \textbf{B1} on the board, Bob might reason that the card is likely to be the \textbf{B2} even if its color is unspecified. Therefore, the hint's implicit meaning becomes ``play this now''. This allows the team to add a card to the board using only one hint token (to identify only rank) rather than two (to identify both rank and color).

Assumptions and the actions they enable are often interchangeably called conventions. They can be explicitly agreed to or emerge naturally between players. They can be used to predict future actions of other players (e.g. ``If I hint \textbf{2} to Bob, he will play the card'') or to reduce the number of different game states a player believes themselves to possibly be in (e.g. in a scenario where Alice had chosen not to give a hint, Bob might reason ``I probably have no playable cards'').

Conventions, either crafted by experts~\cite{cox2015make,bouzy2017playing} or implicitly enacted by a learned policy~\cite{bard2020hanabi,foerster2019bayesian,hu2019simplified} have been used to create agents with near-perfect scores, as long as the entire team shares the same convention or policy. However, these agents tend to be brittle and to fail in uncoordinated teams.

To see why relying on the wrong assumptions can lead to poor results, consider if Alice had been giving hints at random in the example above: this would make the interpretation that the card is playable unlikely to be correct, and playing it would likely lead to a mistake.

A human player in this position might adjust their strategy: if Bob played his third card and it turned out not to be playable, he might stop assuming Alice is giving actionable hints and start waiting until he knows both the color and rank of his cards before playing them in the future.

Intuitively, Bob first hypothesizes a set of strategies Alice could be using (e.g. ``actionable hints'' or ``random hints''), then adjusts hypotheses based on available evidence (the card being unplayable is evidence against the ``actionable hints'' hypothesis) and finally to select an appropriate response strategy (e.g. ``only play cards with full information''). 

This is the intuition behind the meta-agent presented in this paper. It uses rule-based agents generated by MAP-Elites both as the set of hypotheses and the set of possible response strategies, and measurable behavioral characteristics (Communicativeness and Information per Play) as evidence for which hypothesis best approximates its partner's true policy.

\section{Related Work}\label{sec:related}

\subsection{Hanabi-Playing agents}\label{sec:related-hanabi}

Many published Hanabi agents fall into one of two broad categories: rule-based agents and reinforcement learning (RL) agents. 

Unless otherwise specified, all agent scores in this section refer to the average self-play score in the 2-player version of the game as reported by the authors.

We define rule-based agents as agents that follow a list of ``rules'' sorted by priority. Each rule has an optional condition and a corresponding action. If the condition is true for a given game state, the agent takes the corresponding action, otherwise it moves on to the next rule in priority. Examples of rules are:

\begin{itemize}
    \item If a card is guaranteed to be playable, play it.
    \item If a card has a probability of being playable over a threshold, play it.
    \item If a card is known to be useless, discard it.
    \item Discard a random card.
    \item If a partner has a playable card, give a new piece of information about that card, favoring rank over color.
\end{itemize}



The earliest agent in this category follows Osawa’s \textit{Self-Recognition} strategy~\cite{osawa2015solving}, which assesses whether a card is likely to be playable or useless by assuming its partner is using the same strategy. It then filters out card combinations that are incompatible with that strategy. Van den Berg~\cite{van2016aspects} uses simulations to determine the best parametrized variants of similar agents. Walton-Rivers \textit{et al.}~\cite{walton2017evaluating} evaluate these and other agents based on their performance when paired with an evaluation pool of seven agents. Out of all the agents in~\cite{walton2017evaluating}, the one with best self-play score in the 2-player version (as evaluated by Canaan \textit{et al.} in~\cite{canaan2020behavioral}) is \textit{Piers}, with a score of 17.31 (out of 25).

The 2018 and 2019 Hanabi competitions~\cite{walton20192018} provided a framework including implementations of all rules used in~\cite{walton2017evaluating}. The competition had both a self-play (“Mirror”) track and a track for playing with agents known only to the organizers (“Mixed”).


Our work is based on an entry to the 2019 competition by Canaan \textit{et al.} which uses MAP-Elites to evolve behaviorally diverse agents represented by sequences of rules~\cite{canaan2019diverse}. The rule set we use contains rules provided in the competition framework as well as rules implemented for a 2018 entry, also by Canaan \textit{et al.}~\cite{canaan2018evolving}. The best self-play agent of the 2019 entry had a score of 20.51.

Eger's \textit{Intentional} agent~\cite{eger2017intentional} and Liang's~\textit{Implicature} agent~\cite{liang2019implicit} could also be described as rule-based under our definition. They are designed according to Grice's maxims of communication~\cite{grice1975logic} and achieve average self-play scores of 17.1 and 18.9, respectively, in their best variants. When paired with humans, Eger reports an average score of 14.99, while visual inspection of Liang's results suggests an average score around 12 points.


The current state-of-the-art for self-play in Hanabi is a hybrid agent by Lerer et al~\cite{lerer2020improving}. It combines a public ``blueprint policy'', that serves as a baseline convention, with a distributed search protocol that makes it possible for all players to deviate from the blueprint policy when it is computationally feasible to do so. The best blueprint policy is the one used by the RL agent called Simplified Action Decoder (\textit{SAD})~\cite{hu2019simplified}. During centralized training, \textit{SAD}'s observation is enhanced with both a ``greedy'' and an ``exploratory'' action to address the fact that the randomness required for exploration during training makes the agent's actions less informative to other players. During evaluation, only the ``greedy'' action is used and \textit{SAD} achieves a score of 24.01 in self-play. The hybrid agent that uses \textit{SAD} as blueprint achieves a score of 24.61.

\textit{SAD}, as well as other RL agents that preceded it~\cite{bard2020hanabi,foerster2019bayesian} all suffer from the problem that their learned policies are extremely brittle, as observed by their authors and also explored in~\cite{canaan2020behavioral}. Many of these agents seem to operate under conventions that consist of arbitrary mappings of a hinted color or rank to a desired action. For example, an agent might interpret any hint towards the color Yellow as a command to play their fourth card. This is similar to conventions used by hat-guessing agents~\cite{cox2015make,bouzy2017playing} which can also achieve average scores above 24 in at least some settings, depending on the number of players and on rule variants.

The \textit{Other Play} training regime~\cite{hu2020other} addresses this brittleness by using known symmetries to randomly re-label, during training, the actions and states observed by the agent. In doing so, it achieves a self-play score of 24.09 and a cross-play score (with similarly trained agents) of 22.49. \textit{SAD} has also been evaluated with human partners, scoring 9.15 and 15.75 in the vanilla and \textit{Other Play} variants respectively.

While $SAD$ and other RL agents achieve higher self-play scores than any rule-based agents, we nonetheless use the latter as the foundation of this work. The main use for these agents in this paper is as a set of cheap models of diverse behavior. Current RL methods give little guarantees of behavioral diversity, while diverse rule-based agents are very cheap to generate with MAP-Elites. Furthermore, the main advantage of RL agents is stronger performance, but this does not necessarily make them the best to model human players, especially novice players. For example, it is simpler to build a rule-based agent that avoids certain variations of the ``play this now'' convention explained in section~\ref{sub:conventions} than to ensure that an RL agent will never learn these conventions. Furthermore, the resulting RL agents would likely not display near-SOTA performances in the first place.


The only agent we are aware of that attempts to directly model and adapt based on partner behavior in Hanabi is \textit{IS-MCTS}, winner of the 2018 and 2019 CIG/CoG competitions~\cite{goodman2019re}. Similar to the work discussed in this paper, IS-MCTS uses Bayesian updates to model the belief that the current ad-hoc partner is using one of a set of previously known strategies, but they estimate the probability that each strategy would produce the observed action using a neural network instead of behavioral features. However, the author does not provide a detailed analysis of how this adaptation procedure affects ad-hoc performance or compare it with a non-adaptive baseline. \rev{}{Another key difference is that the meta-agent described in this paper adapts by explicitly switching its own policy, while IS-MCTS uses a model of its partner's behavior to predict partner moves during tree search.}

\subsection{Ad-Hoc Team Play and Zero-Shot Coordination}\label{sub:ad-hoc teamplay}

There are two distinct but related theoretical frameworks that attempt to formalize the notion of cooperation we are interested in in this paper.

The first is \textit{Ad-Hoc Team Play}, introduced by Stone \textit{et al}~\cite{stone2010ad}. The goal is to design an agent that is able to cooperate with a team of arbitrary partners, with no prior coordination. The authors state that a good Ad-Hoc team player should be capable of
``assessing the capabilities of other agents, especially in relation to its own capabilities''. The focus is on modelling and adapting to other players. Episode lengths are assumed to be long enough that changing one's policy to account for the characteristics of team mates is possible, such as in a match of soccer.

The second is \textit{Zero-Shot Coordination}, introduced with the \textit{Other Play} agent mentioned in section~{\ref{sec:related-hanabi}}{}~\cite{hu2020other}. The goal is to maximize score when playing a single episode with agents that were independently trained for the same task. \rev{Episodes are assumed short enough that no adaptation within an episode is possible.}{} The focus is on developing policies that are, according to the authors, ``maximally robust to partners breaking symmetries in different ways''.{}

Our work is more aligned with \textit{Ad-Hoc Team Play}, focusing on identifying and adapting to unknown partners. The original framing of the problem can be stated as such: given a pool of potential team mates $A$ and a domain of potential tasks $D$, create an agent $a$ that maximizes the expected payoff when asked to perform a randomly sampled task $d \in D$, paired with a randomly sampled subset of teammates $B \subset A$.

It is important to consider that, while the set of current teammates $B$ is not known to the \textit{agent} at the start of each episode, the set of potential teammates $A$ may or may not be known to the \textit{designer} of an agent. In the 2018 and 2019 Hanabi competitions, for example, $A$ included a mix of hand-crafted and evolved agents, which were not known to participants.

If $A$ is unknown, a designer may imbue an agent with implicit or explicit priors that reflect the designer's beliefs (hypotheses) about $A$. In this paper, we make an explicit distinction between the designer's hypotheses $H$, from which we sample partners during training, and the actual evaluation pool $E$, from which we sample partners during evaluation.

In general, $H$ and $E$ could be arbitrary distributions sampled using arbitrary strategies, but in our experiments we assume them to be discrete ``pools'' from which we sample uniformly for simplicity. In section~\ref{sec:overview}, we also introduce a third pool, $R$, corresponding to a set of discrete strategies the agent can choose to employ at any point during evaluation.

\subsection{Quality Diversity and MAP-Elites}\label{sub: QD and MAP-Elites}

Quality Diversity~\cite{pugh2016quality} (QD) algorithms are a class of population-based search algorithms that aim to generate a large number of solutions that are behaviorally diverse and of high quality. Behaviorally diverse means the agents are distributed representatively across a behavior space induced by one or more behavioral metrics. High quality means the agent performs well according to some fitness function.

Diversity of behavior can be pursued either as a desirable target in its own right or as an intermediate step to high-quality solutions in deceptive fitness landscapes, as showcased by novelty search~\cite{lehman2011abandoning}. QD differs from novelty search, however, because it does not optimize for novelty alone, but searches for both behavioral diversity and high fitness at once. QD also differs from Multi-Objective Optimization~\cite{deb2014multi}, which searches for trade-offs between one or more objectives, because QD actively attempts to find high-quality solutions in all regions of the behavior space, not just the regions with good trade-offs.

MAP-Elites~\cite{mouret2015illuminating} is an example of QD algorithm that attempts to \enquote{illuminate} the behavior space by mapping each individual to a behavioral \enquote{niche}, while maintaining an archive of the best individual (an elite) in each niche. MAP-Elites was first proposed to pre-compute a variety of effective gaits for a six-legged robot so that, when the robot suffers damage, it can quickly search for a gait that adapts to the damage and allows it to keep moving at a decent pace~\cite{cully2015robots}.

In this paper, we use MAP-Elites to generate the pools $E$, $H$ and $R$. The choice of a QD algorithm for the generation of these pools represents the desire that the evaluation experiments test for a wide range of skills (for $E$), that the agent is able to model a varied set of ad-hoc partners (for $H$) and that the agent is able to select from a varied set of response strategies (for $R$).

\rev{}{
Our implementation of MAP-Elites (described in section}~\ref{sec:pre-training-map-elites})\rev{}{ is relatively straight-forward, but the reader might also be interested in recent improvements such as MAP-Elites with sliding boundaries}~\cite{fontaine2019mapping}\rev{}{, which takes into account the density of solutions in behavior space when drawing the boundaries between niches, and Covariance Matrix Adaptation MAP-Elites (CMA-ME)}~\cite{fontaine2020covariance}\rev{}{, which uses an adaptation mechanism inspired by CMA-ES}~\cite{hansen2016cma}\rev{}{  for increased performance.}

\subsection{Bayesian Inference}

There is a long history of using Bayesian Inference to model the behavior of other agents, such as the Bayes formulation of Hyper-Q learning~\cite{tesauro2003extending} and Interactive Bayesian Reinforcement Learning~\cite{chalkiadakis2003coordination, hoang2013general}. While these may provide theoretical frameworks and some guarantees of convergence or optimality, they are usually evaluated using smaller environments such as Rock-Paper-Scissors or small grid-worlds.

More recently, a method for Bayesian delegation of tasks in a version of the cooperative real-time game \textit{Overcooked} was proposed by Wang \textit{et al.}~\cite{wang2020too}. It uses Bayesian inference to infer which sub-task the other player currently intends to perform. In contrast, we use Bayesian Inference in this paper to infer which class of agent (represented by one of several known strategies) the other player belongs to.

\textit{MeLIBA} (Meta Learning Interactive Bayesian Agents)~\cite{zintgraf2021deep} is a method used to maintain beliefs over partner strategies with similarities to our work. It uses the latent variables of a hierarchical variational autoencoder to model beliefs over agents in a treasure hunt task on a grid world. While their method allows the modelling of non-stationary agents during a single match, our approach assumes each partner comes from a fixed pool of potential partners, but allows for adaptation in-between matches.

As will be detailed in section~\ref{sub:bayes-update}, our adaptive step can be seen as a Gaussian Naive Bayes Classifier~\cite{rish2001empirical} that uses the pool $H$ as classes. However, the Gaussian and naive independence assumptions are not intrinsic to the method, and the distribution for each dimension could be independently modeled given a richer record of the training data.

\section{Behavioral Features}

\rev{}{
Our goal is for the meta-agent to play well with partners that exhibit diverse behavior. In this section, we present the behavioral features used to characterize an agent's behavior. These are the metrics used in section}~\ref{sec:pre-training-map-elites}\rev{}{ to generate agents with MAP-Elites and also used by the meta-agent itself in section}~\ref{sub:ad-hoc adaptation meta agent}\rev{}{ to estimate an unknown partner's identity. 

We also examine some previously published agents}~\cite{osawa2015solving,van2016aspects,walton2017evaluating,bard2020hanabi,canaan2020behavioral} \rev{}{in light of these behavioral features, investigating how they behave with respect to our chosen behavioral features.}

\subsection{Definition of Behavioral Features}\label{sub:definition-behavioral-features}

We defined the following behavioral features for Hanabi:
\begin{itemize}
    \item \textbf{Information per Play (IPP):} whenever an agent plays a card, we verify whether it knows its color and/or rank. Each of these is considered one piece of information. The verification takes into account both positive and negative facts implied by hints (e.g. a hint of Red implies all other cards to be non-Red) but, for simplification, does not take into account facts that can be deduced through a process of elimination by looking at the discard pile and other players' hands. 
    
    We count how many pieces of information the agent knows (either 0, 1 or 2) for each card that is played, then average this value across all played cards. Finally, we divide by 2 to get a number between 0 and 1. An agent scoring 1 in this dimension only plays cards that are fully known (both color and rank), whereas an agent scoring zero would only play cards it knows nothing about.
    
    \item \textbf{Communicativeness:} defined as the fraction of time an agent will choose to give a hint if a hint token is available at the start of the turn. An agent scoring 1 in this dimension would always give a hint if possible, being fully communicative, while an agent scoring 0 would never give any hints.

\end{itemize}

These features were chosen because they are easy to measure and we believe that they are strategically meaningful. In particular, IPP was meant as proxy for a pattern that can be observed in game-play between humans: inexperienced players usually only play cards they know everything about (which implies IPP close to 1), where more experienced players are more often comfortable playing cards under partial information as a result of either accounting for other player's apparent beliefs and intentions (theory of mind) or of following a particular pre-established convention.

IPP is a successor to the Risk Aversion feature used in~\cite{canaan2019diverse}. Risk Aversion reflects the average probability that a card is playable, from the perspective of the agent, over all played cards. However, Risk Aversion is a hard metric to estimate during gameplay because it relies on hidden information: the probability that a card is playable from an agent's perspective depends on the cards it sees in the hands of its partners, which is not known to the partners.

Agents with low IPP also have more intuitive behavior than those with low Risk aversion: an agent with low IPP simply plays cards of which little or no information is known (possibly as result of a convention), whereas an agent with low Risk aversion is defined as one that only plays cards that are known to be very unlikely to be playable. Such an agent could never achieve high scores.

We suspected that the highest-scoring behavior in self-play would fall at some value much greater than 0, but lower than 1 for both dimensions: a 0 in either dimension leads to obviously degenerate play, but good play likely requires playing cards under some uncertainty (implying IPP $<$ 1) and sometimes passing up the opportunity to give a hint so that the other player can better utilize the hint token (implying Communicativeness $<$ 1). 

Note also that, while these dimensions help describe an agent's play, they don't completely determine it. Communicativeness does not tell us \textit{which} hint will be given, only the likelihood that \textit{some} hint will be given if a hint token is available. Similarly, IPP does not tell us \textit{whether} the agent will play a card, only how much is known on average about it given that it was played.

Each metric takes values in the range of [0,1], and we discretize them for this paper at intervals of 0.2, defining 5 intervals in each dimension of the behavior space. This amounts to a total of 25 niches, in contrast to the 20 by 20 discretization used in~\cite{canaan2019diverse}, which amounted to 400.

The smaller number of niches is due to the fact that the number of match-ups between $H$ and $R$ scales quadratically when $H=R$. This reduces the computational complexity of the offline training step and also makes the behavior of the meta-agent easier to manually inspect.

\subsection{Evaluation of Behavioral Characteristics of Existing Agents}

Before discussing how the agents generated by MAP-Elites are used by the meta-agent, it is worth looking at how Communicativeness and IPP, the behavioral characteristics chosen for use in MAP-Elites, can be used to analyse existing agents in the literature.

To do this, we first looked at six rule-based agents from the Hanabi literature: \textit{Internal} and \textit{Outer} by Osawa~\cite{osawa2015solving}, Van den Berg's best-performing agent from~\cite{vinyals2019grandmaster} (which we refer to by \textit{VDB}), and \textit{Flawed}, {IGGI} and \textit{Piers} by Walton-Rivers \textit{et al.} We used the re-implemented versions of these agents by Canaan \textit{et al.} in~\cite{canaan2020behavioral}. We paired each of these agents with each other for 1000 games per pair, measuring the Communicativeness and IPP displayed by each agent in each paired match-up.

Tables~\ref{table:rule-based communicativeness}  and~\ref{table:rule-based IPP} show, respectively, the Communicativeness and IPP values resulting from this evaluation. One of the things this evaluation allows us to see is the extreme degeneracy of Flawed's behavior: when playing with itself, it is the only agent with both Communicativeness (0.06) and IPP (0.04) close to zero. When playing with other agents, its partners also exhibit uncharacteristic behavior. For example, IGGI exhibits IPP $>$ 0.9 with all partners, except when playing with Flawed, in which case IGG exhibits IPP of only 0.68.

Ignoring the match-ups involving Flawed, Communicativeness varied over a wide range, from 0.36 to 0.91, while IPP was relatively high across the board, varying from 0.73 to 0.98. 

\begin{table*}
\caption{Communicativeness of six rule-based agents from~\cite{canaan2020behavioral} playing among themselves. Lines represent the agent being evaluated and columns represent each of their partners (e.g. when IGGI plays with Internal, IGGI displays Communicativeness of 0.36)}
\begin{tabular}{ccccccccccc}
 & IGGIAgent & InternalAgent & OuterAgent & VanDenBerghAgent & FlawedAgent & PiersAgent & Self & Min & Max & Average \\
IGGIAgent & 0.50 & 0.36 & 0.41 & 0.38 & 0.46 & 0.42 & 0.51 & 0.36 & 0.51 & 0.43 \\
InternalAgent & 0.89 & 0.88 & 0.83 & 0.90 & 0.99 & 0.87 & 0.88 & 0.83 & 0.99 & 0.89 \\
OuterAgent & 0.89 & 0.89 & 0.84 & 0.91 & 0.99 & 0.85 & 0.84 & 0.84 & 0.99 & 0.89 \\
VanDenBerghAgent & 0.63 & 0.36 & 0.36 & 0.50 & 0.52 & 0.53 & 0.50 & 0.36 & 0.63 & 0.48 \\
FlawedAgent & 0.28 & 0.17 & 0.17 & 0.36 & 0.06 & 0.26 & 0.08 & 0.06 & 0.36 & 0.20 \\
PiersAgent & 0.64 & 0.47 & 0.56 & 0.53 & 0.50 & 0.58 & 0.58 & 0.47 & 0.64 & 0.55 \\
\end{tabular}
\label{table:rule-based communicativeness}
\end{table*}

\begin{table*}
\caption{IPP of six rule-based agents from~\cite{canaan2020behavioral} playing among themselves. Lines represent the agent being evaluated and columns represent each of their parnters (e.g. when IGGI plays with Internal, IGGI displays IPP of 0.98)}
\begin{tabular}{ccccccccccc}
 & IGGIAgent & InternalAgent & OuterAgent & VanDenBerghAgent & FlawedAgent & PiersAgent & Self & Min & Max & Average \\
IGGIAgent & 0.94 & 0.98 & 0.97 & 0.95 & 0.68 & 0.95 & 0.94 & 0.68 & 0.98 & 0.92 \\
InternalAgent & 0.92 & 0.96 & 0.94 & 0.92 & 0.94 & 0.93 & 0.95 & 0.92 & 0.96 & 0.94 \\
OuterAgent & 0.95 & 0.96 & 0.96 & 0.94 & 0.96 & 0.95 & 0.96 & 0.94 & 0.96 & 0.95 \\
VanDenBerghAgent & 0.77 & 0.81 & 0.79 & 0.78 & 0.69 & 0.80 & 0.79 & 0.69 & 0.81 & 0.77 \\
FlawedAgent & 0.45 & 0.41 & 0.47 & 0.46 & 0.04 & 0.45 & 0.04 & 0.04 & 0.47 & 0.33 \\
PiersAgent & 0.73 & 0.74 & 0.74 & 0.77 & 0.77 & 0.78 & 0.78 & 0.73 & 0.78 & 0.76 \\
\end{tabular}
\label{table:rule-based IPP}
\end{table*}

\section{Overview of the Hanabi Bayesian Meta-Agent}\label{sec:overview}

Our method can be summarized in three phases: 

\begin{enumerate}
    \item \textbf{Pre-training}, where we generate a population of agents to serve as a pool of hypotheses of partner behavior $H$ and a (possibly distinct) evaluation pool $E$.
    \item \textbf{Offline training}, where we use $H$ to compute a pool of response strategies $R$ and a set of identifying information $I(r, h)$, representing the observed features in each match-up between $r\in R$ and $h \in H$ 
    \item \textbf{Online ad-hoc evaluation}, where we use Bayesian inference to maintain a belief distribution about an unknown ad-hoc partner for a number of games. This allows us to then choose actions according to the response from $R$ that is expected to maximize score when paired with agents according to the belief distribution.
\end{enumerate}

Ties are broken arbitrarily in step 3, but this is a rare occurrence since both the belief distribution and the expected match-up scores are real-valued.

Because we use Bayesian inference to choose which agent from the response pool to ``impersonate'', we call our approach a Hanabi Bayesian Meta-Agent. This means the meta-agent does not among possible actions directly, but among response strategies, and merely plays the same action that the chosen response strategy would play at a given game state.

We now provide an overview of each of these phases. Note that the steps taken at each phase are fairly independent from each other, so they can be thought of as modules. For example, we could have used distinct algorithms to generate $H$, $R$ and $E$ or arbitrary behavioral features to compute $I$. The implementation used for the experiments of this paper can be found in our public github repository~\footnote{https://github.com/rocanaan/Hanabi-Map-Elites}. All experiments were conducted in the 2-player version of the game.

\subsection{Pre-Training Phase}

This phase consists of the generation of agents to constitute the pool of hypotheses $H$, which is known to the meta-agent, and the evaluation pool $E$, which might not be. In principle, any technique for generating Hanabi agents could be used. We expect behavioral diversity in the training population to be instrumental when attempting to adapt outside the training population. For this reason, we use MAP-Elites, a Quality Diversity algorithm, to generate populations of training agents using a similar procedure as~\cite{canaan2019diverse}.

\subsection{Offline Training Phase}

This phase consists of two steps:

\begin{itemize}
    \item \textbf{Generation of a response pool $R$}. This is the set of policies the meta-agent can choose from when taking an action. Generally, any technique for generating agents that play well with agents in $H$ or subsets of $H$ could be used, including RL, evolution and tree search. For simplicity, however, we re-use the pool of hypotheses itself as response pool. In other words, $H=R$ for all our experiments. 
    \item \textbf{Identifying information of each match-up $I$.} We call a set of games played by two agents a match-up. Given a match-up $(h,r)$, where $h\in H$ and $r \in R$, we call $I(r,h)$ the set of identifying information of that match-up. In principle, the whole game history or any number of features derived from this history could be used, but to reduce the storage requirements, we store only the average Communicativeness and IPP displayed by $h$ in the match-up and the average score of the match-up. These are the same features used during the generation of agents with MAP-Elites.
\end{itemize}


\subsection{Online Ad-Hoc Evaluation}

During this phase, the agent is paired with a partner $e$ sampled from an evaluation pool $E$ for a short episodes of 10 matches each. During each episode, the meta-agent starts with no information about $e$ and assumes it could be any agent from the pool of hypotheses $H$ with equal probability. This belief distribution is then updated as the meta-agent gains more information across the 10 games of the episode.

On each of the meta-agent's turn, it needs to do two things:

\begin{enumerate}
    \item \textbf{Update the belief distribution.} It does this by keeping track of average of its partner's behavioral characteristics (Communicativeness and IPP) and performing a Bayesian update based on the meta-information stored for that match-up.
    \item \textbf{Select an action.} It does this by performing the action that would be chosen by the strategy in $R$ that maximizes the expected score over $H$, weighted by the belief distribution.
\end{enumerate}

\section{Pre-Training: MAP-Elites}\label{sec:pre-training-map-elites}

In this section, we describe MAP-Elites, which we use to generate the hypotheses. $H$, response pool $R$ and evaluation pool $E$. Over the course of the algorithm, agents are represented by sequences of rules taken from a rule set consisting of rules provided by the CIG/CoG competition framework~\cite{walton20192018} and a 2018 competition entry~\cite{canaan2018evolving}. For domains where a similar rule set is not available, one would have to be created or a different representation agents would have to be used, such as the weights of a neural network.

In MAP-Elites, we first define one or more quantitative or categorical \textit{behavioral characteristics} or \textit{features} that can be used to describe the behavior of a solution. These can be thought of as coordinates in a multi-dimensional \textit{behavior space}. A process of discretization partitions the space into \textit{niches} where all agents have similar behavioral characteristics. For the experiments in this paper, we used the behavioral features defined in section~\ref{sub:definition-behavioral-features} to define the feature space, which was partitioned in intervals of 0.2 for each feature. Since both feature values range from 0 to 1, these results on 5 partitions in each dimension, for a total of 25 niches.

We then conduct an evolutionary process where new candidate solutions are generated as variations (mutation and/or crossover) of existing solutions. Each candidate is assessed for its behavioral characteristics, which allows it to be placed within a niche. If that niche is currently empty, the candidate occupies it, becoming the \textit{elite} of the niche. Otherwise, the candidate's fitness is compared to that of the current elite. The winner becomes the new elite, while the loser is discarded.

As result, MAP-Elites simultaneously attempts to find valid solutions for currently empty niches in its archive and higher-fitness solutions to niches that are already occupied.

\subsection{Representation and operators}~\label{sub:represenation and operators}

We use a similar representation of individuals as the one proposed by Canaan et al~\cite{canaan2018evolving}. Each individual is represented by a chromosome defined by a sequence of 15 integers, where each integer represents one of 135 possible rules which were either initially provided in the Hanabi competition framework~\cite{walton20192018} or implemented for the entry.

An agent's action is determined by simply moving through the rules in the order they appear in the chromosome. The agent outputs the action returned by the first applicable rule. An agent might have rules that never trigger during gameplay (for example, a rule that says \enquote{discard a random card} would never trigger if it comes after \enquote{discard your oldest card}). An agent can also have duplicate rules, in which case the second instance of the rule will never trigger (assuming the rule either triggers or not deterministically, which is true for the rules we are using). Nevertheless, these unused or repeated rules are part of an agent's genetic representation and can be passed on to its offspring. We selected 15 as chromosome length because our agents from~\cite{canaan2018evolving} rarely had more than 10 different rules activated. 

The first few chromosomes (in our experiments, $10^4$) are implemented by sampling rules uniformly at random from the ruleset, while the remaining chromosomes are generated by mutation and crossover of the elite in a random niche. Mutation is implemented by randomly replacing each rule in a chromosome with a random new rule with probability 0.1. Crossover happens with probability 0.5 and is implemented by selecting another individual from the population and randomly selecting (with probability 0.5) the corresponding rule from either parent at each gene. 

\subsection{Pseudocode of the MAP-Elites algorithm}

With these metrics, representation and operators in mind, algorithm~\ref{algorithm:MAP-Elites} shows the abstracted pseudocode of the MAP-Elites algorithm.

\begin{algorithm}\label{algorithm:MAP-Elites}

\KwResult{$A$, an archive with each niche's elite.}
 $generation \gets 0$\;
 $A \gets \emptyset$\;
 \While{$generation < G$}{
  $c \gets newChromosome(A, generation)$ \;
  $x \gets makeAgent(c)$\;
  $f \gets fitness(x)$\;
  $i,j \gets niches(x)$\;
  \eIf{$A_{i,j} = null$ }{
  $A_{i,j} \gets c$;
  }
  { 
     $f_{elite} \gets fitness(makeAgent(E_{i,j}))$ \;
     \If{$f > f_{elite}$}{
     $A_{i,j} \gets c$
     }
  }
 }
 \caption{MAP-Elites}
\end{algorithm}

In our experiments, the functions \textit{newChromosome}, \textit{makeAgent}, \textit{niches} are implemented as described below:

If \textit{generation} $<10^4$, \textit{newChromosome} returns a new list of 15 rules by sampling the rule-set uniformly. Otherwise, the new chromosome is generated by mutation and crossover of random parents sampled from $A$, as described in section~\ref{sub:represenation and operators}.

\textit{makeAgent} simply returns an agent instance that follows a policy determined by applying the chromosome rules in order, as also described in section~\ref{sub:represenation and operators}.

\textit{fitness} returns the average score of the agent after playing 100 matches in self-play mode. While these matches are played, a number of statistics can be recorded, such as the number of hints given, the number of turns where a hint token was available, the total number of cards played and how many pieces of information was known about each played card. 

\textit{niches} calculates the Communicativeness and IPP values of a candidate based on these stored statistics, and converts these behavior features (which take values between 0 and 1) into two integer indexes, according to the discretization of the behavior dimensions. In our case, a value between 0 and 0.2 corresponds to the first niche on a dimension, a value between 0.2 and 0.4 to the second niche etc, until the last niche corresponding to values between 0.8 and 1. 

After that, the program checks whether an elite has already been assigned to the corresponding entry $A_{i,j}$. If that entry is empty, the chromosome of the current agent is stored in that cell. Otherwise, we re-calculate the fitness of the elite in $A_{i,j}$, using the same seed as used for the candidate, and keep the agent with the best agent in the niche.

We recalculate the elite's fitness rather relying on a stored value due to the stochastic nature of the fitness evaluation, to avoid a single overestimation of fitness to result in an elite that is exceedingly hard to replace.

\section{MAP-Elites Results}\label{sec:map-elites-results}

We used Map-Elites to generate and evaluate agents by executing three separate runs of algorithm~\ref{algorithm:MAP-Elites}. Each run generated and evaluated a total of $10^6$ candidate individuals and recorded the chromosome of the elite in each of the 25 behavioral niches where a score greater than zero was achieved. A run's \textit{coverage} is defined as the number of niches successfully filled by an elite in the run.

At the end of each run, we re-evaluated the final elite of each niche by playing 1000 self-play games. We also evaluated each agent's average performance when paired uniformly with all agents from that run (including itself), which we call the agent's pairwise score. Table~\ref{table:runs} shows each run's coverage, the maximum self-play score of any elite in each run during this re-evaluation, the average self-play score of agents in all covered niches, the average pairwise score of agents in all covered niches, the correlation (Pearson coefficient) between all agent's self-play and pairwise performances..

\begin{table*}[]
\centering
\caption{Coverage, max self-play score, average self-play score, average pairwise score and correlation (Pearson coefficient) between self-play and  pairwise score of agents in the three MAP-Elites runs. Average scores take into account covered niches only.}
\begin{tabular}{|c|c|c|c|c|c|}
\hline
\textbf{Run} & \textbf{Coverage} & \textbf{Max Self-Play Score} & \textbf{Average Self-Play Score} &  \textbf{Average Pairwise Score} & \textbf{Correlation Self-Play / Pairwise} \\
\hline
Run 1 & 22 & 19.54 & 11.36 &  8.71 & 0.92 \\
\hline
Run 2 & 22 & 19.95 & 11.66 &  8.14 & 0.97 \\
\hline
Run 3 & 22 & 20.00 & 11.59 &  9.52 & 0.91 \\
\hline
\end{tabular}
\label{table:runs}
\end{table*}


\begin{figure*}
\centering
\includegraphics[scale=0.25]{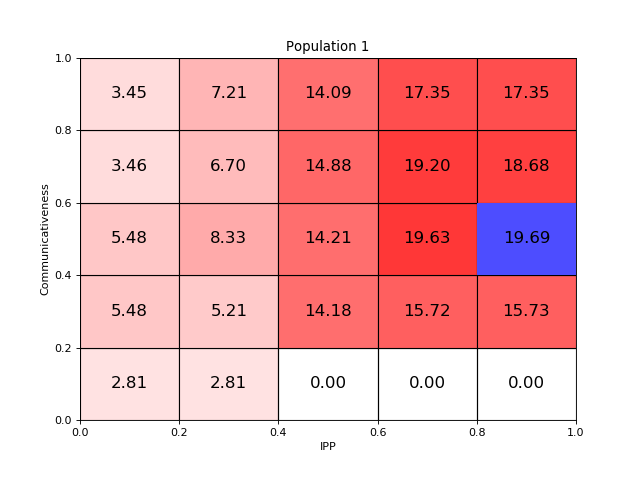}
\includegraphics[scale=0.25]{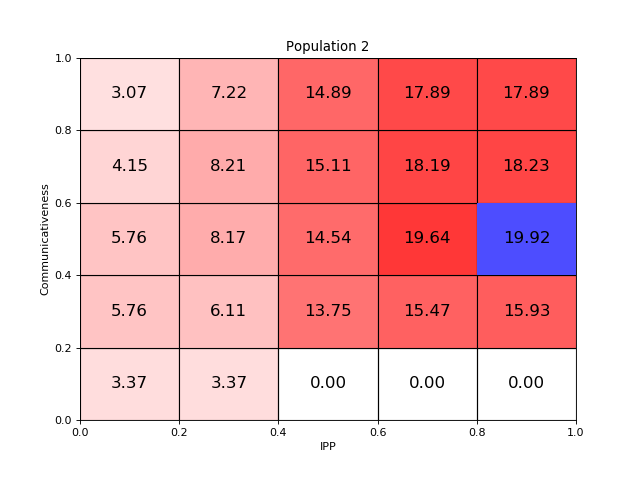}
\includegraphics[scale=0.25]{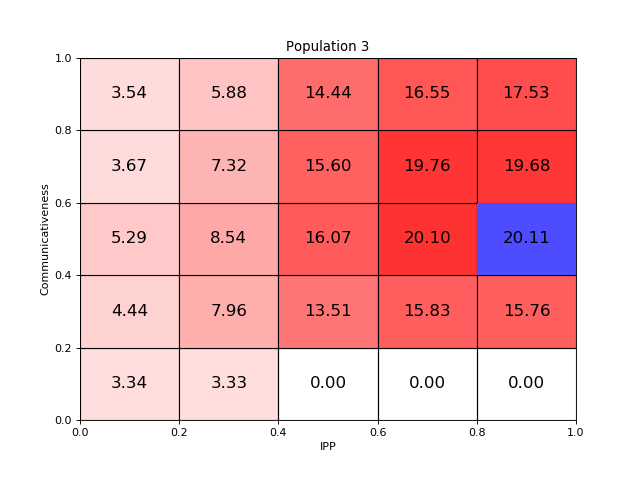}
\caption{Main results of the MAP-Elites experiment on a 5 by 5 grid after reevaluating each elite in each run for 1000 games each. Values represent the fitness (score) of the best individual in that niche, with redder entries corresponding to higher scores. The maximum score for each run is highlighted in blue.}
\label{fig:1Mresults}
\end{figure*}

The coverage of all runs was 22 out of 25 possible niches. The best max score and average score varied slightly from run to run, ranging from 19.69 to 20.11 (max score) and 11.44 to 11.74 (average score).

Figure~\ref{fig:1Mresults} shows the fitness (self-play performance) of the elite in the 25 niches of each run. The three runs showed a very similar fitness landscape, with highest scores concentrated in the region with high values of both Communicativeness and IPP (upper right of each graph). Three niches in the bottom right of the graph (low Communicativeness, high IPP) were not filled in any of the runs. An agent in this region would rarely give hints, yet only play cards it knows a lot of information about. In self-play, such an agent would never give enough hints to its partner to satisfy their high IPP requirement and would thus never play any cards and score zero as result.

Manual analysis of the chromosomes suggests that agents with high Communicativeness tend to have many ``Hint'' rules near the top (highest priority) of their chromosome, while agents with low Communicativeness tend to have them at low prioirty. Agents with high IPP tend to favor cautious ``Play'' rules, agents with intermediate IPP tend to play the most recently hinted card or require an intermediate probability of success before playing a card, and agents with low IPP tend to play cards at random or with a very low probability requirement. 

The high Communicativeness and low IPP region is somewhat degenerate: it requires an agent to give frequent hints, but nonetheless play cards it knows nothing (or little) about. Some agents near the top left of the map achieve this by having a high-priority rule that plays a card essentially at random, but only if the team has three lives. This usually results in playing an unknown card and losing a life, at which point the rule can no longer fire. Other ``Play'' rules, if present, are low-priority, resulting in low IPP. The remaining rules are mostly ``Hint'' rules, resulting in high Communicativeness.

A complete list of the chromosomes all individuals and their rules can be found within our repository under Analysis.

\begin{figure}
\centering
\includegraphics[scale=0.35]{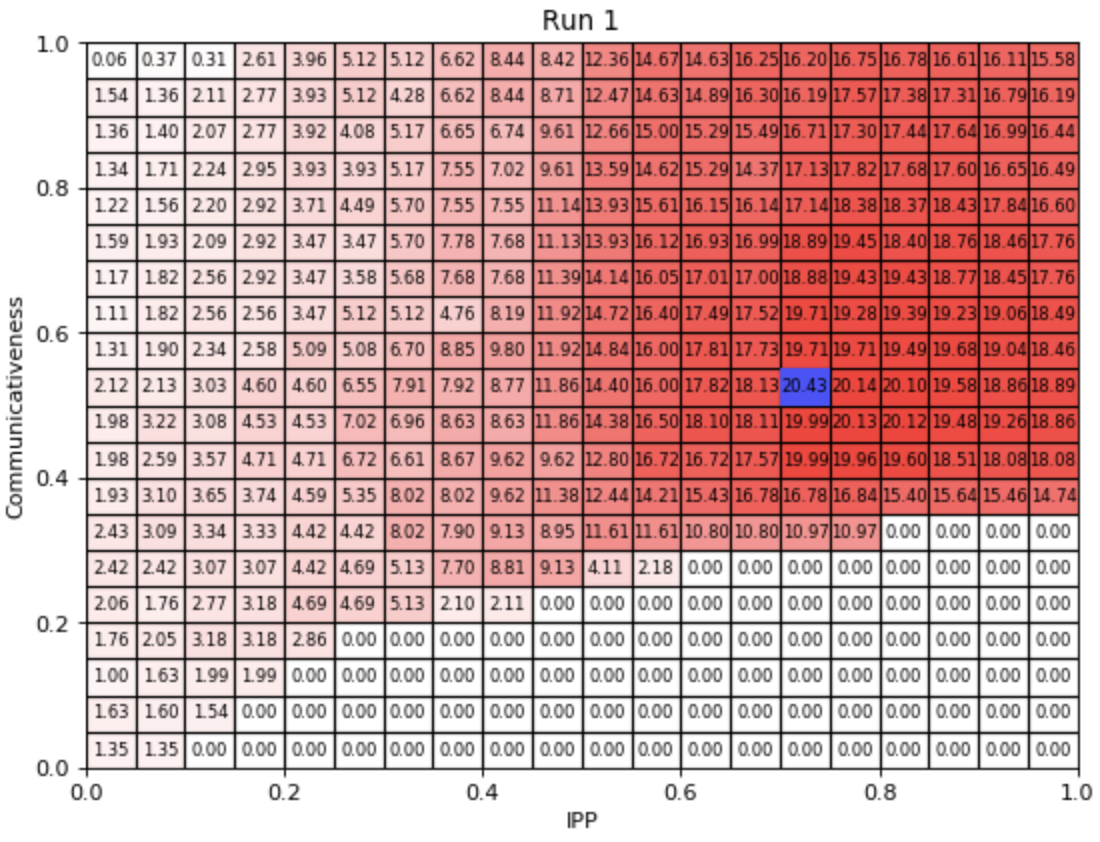}
\caption{Results of population 1 from the original experiments on a 20 by 20 grid, taken from~\cite{canaan2019diverse}. This higher-granularity view shows that the best self-play scores tend to be produced at intermediate communicative levels and high, but not extreme, IPP levels. }
\label{fig:map400}
\end{figure}

The best agent in all three runs occupied the niche with IPP ranging from 0.8 to 1 and Communicativeness from 0.4 to 0.6. Manual inspection of these agents revealed that all three actually have IPP values that would place them right at the leftmost boundary of their niches, within a margin of 0.01 from the boundary at 0.8. Figure~\ref{fig:map400} shows the result of the first run of the corresponding experiment in~\cite{canaan2019diverse}, with finer granularity. Comparing the two versions of the experiment, we see that, in both versions, the best-performing agents occupy the region with intermediate values of Communicativeness (around 0.5) and high, but not extreme, values of IPP (around 0.7 to 0.9).

Regarding pairwise performance, we see that the average pairing between two agents from the same run yields an average score between 8.14 (run 2) and 9.52 (run 3). This is much worse than the average self-play score of these agents (around 11 points), meaning they play much worse with one another than they play with themselves. However, an agent's pairwise performance is strongly correlated with their own self-play scores (Pearson coefficient of 0.91 to 0.97), meaning that for the most part, the better an agent is at self-play, the better we should expect it to perform with other agents. In the original paper~\cite{canaan2019diverse}, we also verified that the farther apart two agents niches were on the map, the worse the two agents tended to perform with one another.

While this analysis aggregates the scores of all match-ups between the agents from each run, we use the score of each individual match-up as the basis for the offline training described in section~\ref{sec:offline-training}.

In the original paper~\cite{canaan2019diverse}, we also assessed the degree of similarity between agents that occupied the same niche in different runs, and how well these agents played with each other. This is relevant to the experiments of this paper where the meta-agent can be trained with one population and evaluated with other: if agents are too similar from run to run, this experimental set-up would not be meaningfully different than using the same population for training and evaluation.

For space considerations, we omit the full details of this analysis, but report the main results: using the populations of the original experiment, the Hamming Distance between the chromosomes of corresponding agents was found to be 14.24 out of a maximum of 15. This means corresponding agents shared, on average, less than 1 out of 15 genes with the same rule at the same position. On a sample of recorded game states, corresponding agents selected the same actions per game state around 60\% of the time. This suggests that agents are similar, but not to a degree where their actions are completely predictable, from run to run.

\section{Offline Training}\label{sec:offline-training}

For the offline training step, the meta-agent receives a population of agents to serve as the pool of hypotheses $H$ and outputs population of agents to serve as a response pool $R$. In this paper, we simply use $R=H$.

It then plays a number of games between each pair of agents $(r,h)$, where $r\in R$ and $h\in H$, and records features of each match-up which constitutes the identifying information $I(r,h)$. During evaluation, $I$ will be used to maintain beliefs over $H$ for each ad-hoc partner, so that an appropriate response from $R$ can be selected.

Eq.~\ref{eq:generalist} shows the definition of a ``generalist'' strategy we can compute in this phase. This is simply the strategy from $R$ that maximizes the expected score when paired with agents sampled uniformly from $H$, or equivalently the one with the highest average score across all match-ups. This lets us establish a useful baseline for the meta-agent, since any agent with smaller score than the generalist would be better off simply following the generalist strategy when paired with $H$.

\begin{equation}\label{eq:generalist}
Generalist(R,H) = \underset{r \in R}\argmax ( \sum_{h\in H}{score(r,h)})
\end{equation}

 Eq.~\ref{eq:oracle} defines an ``oracle'' strategy for each match-up representing which strategy the meta-agent should pick if it knew its partner. If $E=R$, this represents the best score we can hope to achieve if we are limited to picking one strategy from $R$ at the start of each game.

\begin{equation}\label{eq:oracle}
Oracle(R,h\in H) = \underset{r \in R}\argmax ( score(r,h))
\end{equation}

Note that the oracle is defined for a given partner $h\in H$ while the generalist is defined for the whole training population $H$.

For our experiments, each agent was trained using one of the three populations generated by MAP-Elites in section~\ref{sec:map-elites-results} as the pool of hypotheses $H$ and one (possibly the same) population used as evaluation pool $E$. While we could have used arbitrary agents as $R$, including agents evolved or trained to maximize score when paired with subsets of $H$, we chose to skip this step and use $R=H$ in all the experiments. In other word, the pool of strategies the meta-agent can choose to enact is the same as the pool of strategies it expects its ad-hoc partners to be using.

For training, we play 400 games between each pair of agents $(r,h)$ and record, for each of these matchups, the average communicativeness and IPP displayed by $H$ as well as the average score of the match-up. After this step, the meta-agent knows what behavioral features it expects from each partner in $H$ given its own response strategy, as well as the expected score of each response strategy.



\section{Ad-Hoc Evaluation: Bayesian Adaptation}\label{sub:ad-hoc adaptation meta agent}

\subsection{Initialization}

Each phase of ad-hoc evaluation is divided in $k$ episodes. For each episode, we play $g$ games between the meta-agent and each partner agent in the evaluation pool $E$. The meta-agent is provided with a consistent ``dummy'' ID for each evaluation partner, which allows it to maintain all relevant information between games of an episode but not to identify which agent in $H$ (if any) a given evaluation partner corresponds to. At the end of each episode, the meta-agent is re-set to its initial state.

When first playing with an evaluation partner $e \in E$, the meta-agent initializes a uniform belief distribution $B_0(e,h)$ which assigns equal probabilities to the hypothesis ``My current ad-hoc partner $e$ uses the same strategy as $h$" for every $h \in H$.

\begin{equation}
    B_0(e,h) = \frac{1}{|H|}, \forall h \in H
\end{equation}

\subsection{Bayesian Update} \label{sub:bayes-update}

The belief distribution is periodically updated upon fixed intervals, based on the number of games or turns since the last update for each evaluation partner. When the agent is required to perform an update for interval $i+1$, it does so using Bayes' rule, given the observation history collected since the previous interval $O_{i,i+1}$ and given the fact that it had been playing according to some strategy $r_{i}$ since the last update:

\begin{equation}
    B_{i+1}(e,h| O_{i,i+1}, r_{i}) \sim P(O_{i,i+1}|r_{i},h)B_i(e,h)
\end{equation}

We break this down as such:

\begin{itemize}
    \item The left-hand side is the posterior and denotes the new belief given the observed history since the last update and the fact that the meta-agent had been using strategy $r_i$.
    \item $P(O_{i,i+1}|r_{i},h)$, is the likelihood term and represents the probability that this history would have been observed in a match-up between $r_{i}$ and $H$. This will be explained below.
    \item $B_i(e,h)$ is the prior belief that $e$ corresponds to $H$. Here we omit the conditional on $r_{i}$ since the prior belief is independent from the choice of strategy.
\end{itemize}

The left-hand side is proportional to the right-hand side and we use a normalization constant (which is the same for all $h \in H$) to ensure all posterior beliefs add up to 1 at the end of the update step.

Before explaining the likelihood term, it is useful to clarify why the posterior and the likelihood depend on the choice of $r_i$. Put simply, the behavior of $e$ depends not only on the policy used by $e$ itself, but also on the strategy of the meta-agent. 

To see why this must be the case, consider a degenerate example: suppose that $e$ is a ``shy'' agent that never uses hint actions until its partner ``breaks the ice'' by using a hint action first. After the ice is broken, $e$ follows some arbitrary policy. The communicativeness displayed by $e$ will depend on $r_i$ in the following way: if $r_i$ is also shy, the communicativeness will be zero. Otherwise, it will depend on $e$'s underlying policy.

In a previous iteration of the meta-agent, we had not conditioned the observed behavior to the response behavior. We had also limited our action selection (see section~\ref{sub:action selection}) to the best response to the belief with highest likelihood (rather than a weighted response). The combination of these factors led to a worse score than the Generalist baseline in that iteration of the meta-agent.

Coming back to the likelihood term, in our experiments we take $O_{i,i+1}$ to be a tuple representing the average observed Communicativeness and IPP of $e$ since the previous update, on match-ups played between $r_i$ and $e$. This average can be computed simply by computing the ratio of hints given to hints that could have been given over the period (Communicativeness) and the normalized ratio between pieces of information known per card played and number of cards played (IPP). We denote this tuple of observed values $OBS$:

\begin{equation}
    OBS = (Comm_{(r_i,e)},IPP_{(r_i,e)}),
\end{equation}

The likelihood represents the probability, for each training partner in $H$, that it would display the observed behavioral characteristics on a match-up with $r_i$. This probability could be estimated directly from training data, but that would require us to maintain a detailed histogram for each match-up and each feature. For simplicity, we chose instead to store only the average Communicativeness and IPP of each match-up, which we denote $\mu_{Comm}$ and $\mu_{IPP}$ respectively. We can then model the two features as if they came from two independent normal distributions with these averages and standard deviation $\sigma = 0.1$. The value 0.1 was chosen empirically based on initial experiments with the meta-agent.

We can finally subtract, for each feature, the expected value (obtained during training) from the observed value. If $D_{Comm}$ and $D_{IPP}$ are the differences along the two dimensions, we have:



\begin{equation}
\begin{split}
    &p_1 = f(D_{Comm},\mu_{Comm},0.1) \\
    &p_2 = f(D_{IPP},\mu_{IPP},0.1) \\
    &p_{joint} = p_1\cdot p_2
\end{split}
\end{equation}

Where $f(x,\mu,\sigma)$ is the p.d.f. of the normal distribution with mean $\mu$ and variance $\sigma^2$ evaluated at $x$. $p_{joint}$ is the estimated probability that the chosen match-up would display the joint observed values under our assumptions. 

As previously mentioned, this iteration of meta-agent effectively acts as a Gaussian Naive Bayes Classifier~\cite{rish2001empirical} but both the Gaussian and naive independence assumptions could be dropped if given a richer record of the training data.

\begin{figure*}
\centering
\includegraphics[scale=0.3]{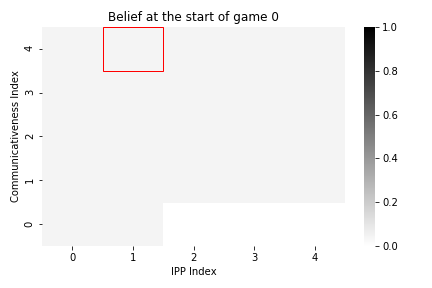}
\includegraphics[scale=0.3]{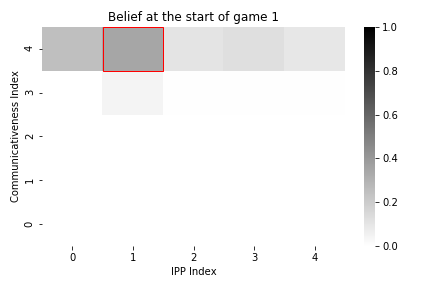}
\includegraphics[scale=0.3]{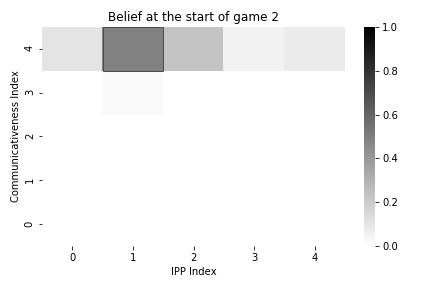}
\includegraphics[scale=0.3]{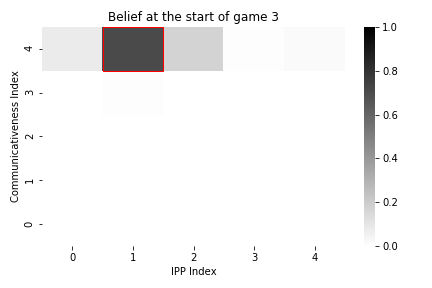}
\includegraphics[scale=0.3]{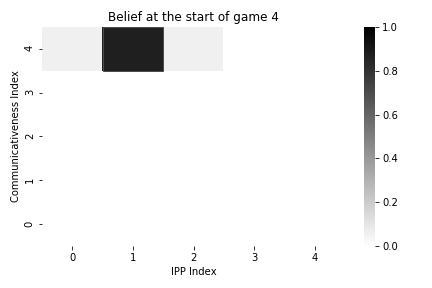}
\includegraphics[scale=0.3]{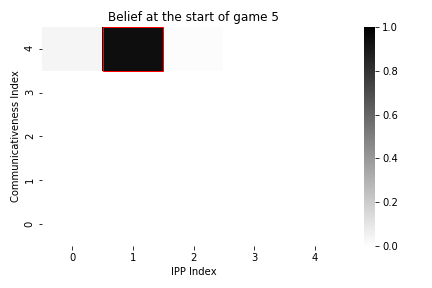}
\caption{Visual representation of progression of the belief distribution of the meta-agent over six-games of a randomly-chosen evaluation match-up. The partner's true identity is unknown to the agent, but highlighted in red for illustration, at the niche with indexes 1 and 4 for IPP and Communicativeness respectively. The meta-agent's belief starts uniformly distributed over all valid training partners (top left). In this example, most of the belief distribution accumulates at the correct identity of the partner (bottom right), but in general, there is no guarantee of convergence.}
\label{fig:beliefs}
\end{figure*}

To illustrate how the belief distribution can vary over the course of a few games, figure~\ref{fig:beliefs} shows a visual representation of the belief distribution over the six first games of a randomly chosen match-up of our evaluation.

\subsection{Action Selection}\label{sub:action selection}

Finally, the meta-agent calculates the response $r^*$ from $R$ that maximizes expected score weighted by the belief distribution $B$ and selects actions using the policy associated with the chosen response.

\begin{equation}
    r^* =  \underset{r \in R}\argmax( \sum_{h\in H}{score(r,h)}\cdot B(e,h))
\end{equation}

\definecolor{mygray}{gray}{0.8}

\section{Ad-Hoc Evaluation Results}

We trained 3 variations of the meta-agent, using each of the 3 MAP-Elites populations as $H$ and $R=H$, for a total of 9 instances of the meta-agent:

\begin{itemize}
    \item An \textbf{Oracle} meta-agent, representing the scenario where the true niche of each evaluation partner is provided by an ``oracle'', removing the need to perform Bayesian updates. While this agent is cheating, it represents an upper bound of the scores we can hope to achieve, given a choice of $H$ and $R$, and assuming $H=E$.
    \item A \textbf{Generalist} meta-agent, which always follows the generalist policy. This is equivalent to setting the adaptation interval to infinite. This serves as our baseline and lower bound, since given a choice of $H$ and $R$ it is always possible to skip the belief updates and simply choose the strategy from $R$ with the highest expected score over $H$.
    \item An \textbf{Adaptive} meta-agent, which performs the Bayesian update at the start of each game. This variant is the main focus of our experiment. Note that the Adaptive meta-agent will always select the same response strategy as the Generalist for the first game. This follows from the definition of the Generalist and from the uniform initialization of the belief distribution.
\end{itemize}

Other adaptation intervals were briefly considered, but longer intervals did not improve performance and shorter intervals (measured in game turns instead of full matches) only moderately degraded performance. While shorter intervals would provide for adaptation within a single game, there is the risk that an agent's average behavior characteristics vary over the course of a game (e.g. an agent might give more hints in the early turns of a game than later on).

In total, we had 3 variations (Oracle, Generalist and Adaptive), each trained using one of the 3 populations as hypothesis pool $H$, then evaluated with using one of the 3 populations as evaluation pool $E$, for a total of 27 scenarios.

Each scenario was further divided in $k=200$ independent episodes where the meta-agent plays $g=10$ games with each agent in $E$. Note that the meta-agent re-uses all of its training information (which we call $I$ in section~\ref{sec:offline-training}) but re-sets its evaluation game history (which we call $O$ in section~\ref{sub:bayes-update}) from episode to episode. 

\begin{table}[]
    \centering
    \caption[Results of the ad-hoc adaptation experiments with the meta-agent]{Results of the ad-hoc adaptation experiments with the meta-agent. Numbers reflect the average score of 2000 games, spread into 200 episodes of 10 games each. The meta-agent is re-set between episodes. Grayed cells represent scenarios where training and evaluation populations were the same.}
    \begin{tabular}{ccccc}
    & & \multicolumn{3}{c}{Evaluation Population}\\
    Training Population & Type of agent & 1 & 2 & 3 \\ \hline
    1 & Oracle & \cellcolor{mygray}13.42 & 12.83 & 13.39 \\
    1 & Adaptive & \cellcolor{mygray}13.16 & 12.58 & 13.49 \\
    1 & Generalist & \cellcolor{mygray}12.94 & 12.49 & 13.67 \\
    1 & Random Response & \cellcolor{mygray}8.85 & 8.48 & 9.17\\ \hline
    2 & Oracle & 12.40 & \cellcolor{mygray}12.80 & 13.04 \\
    2 & Adaptive & 12.32 & \cellcolor{mygray}12.23 & 13.07 \\
    2 & Generalist & 12.52 & \cellcolor{mygray}11.90 & 13.11 \\
    2 & Random Response & 8.27 & \cellcolor{mygray}8.02 & 8.75 \\\hline
    3 & Oracle & 13.18 & 12.50 & \cellcolor{mygray}14.00 \\
    3 & Adaptive & 13.04 & 12.35 & \cellcolor{mygray}13.99 \\
    3 & Generalist & 13.06 & 12.41 & \cellcolor{mygray}13.91 \\
    3 & Random Response & 9.02 & 8.81 & \cellcolor{mygray}9.58 \\
    \end{tabular}
    \label{table:bayes_results}
\end{table}


Table~\ref{table:bayes_results} summarizes the results of these experiments. The gray cells represent ``in-distribution'' scenarios where the training and evaluation populations are the same. These scenarios are meant to test whether the Adaptive meta-agent is able to identify its training partners by observing their behavioral features over the course of the 10 games of each episode. If the agent is able to do so, its performance should fall between the Oracle's and the Generalist's, since the Oracle represents the case where the partner is immediately successfully identified, and the Generalist represents a strategy that is (within $R$) best on average with all partners, but not necessarily optimal for any one partner. A lower score than the Generalist would mean that the meta-agent took adaptive steps that moved it, on average, from the Generalist strategy (which is always chosen for the first game of each episode) to a worse response strategy.

    After 200 episodes, we see that the Adaptive version's average score was below the Oracle's, but above the Generalist's. We performed Welch's t-test~\cite{welch1947generalization} between the distribution of episode scores of the Oracle and Adaptive versions; and between the Adaptive and General versions of the agent. For \rev{populations 1 and 2}{training population 1 evaluated with population 1 and for training population 2 evaluated with population 2}, it rejects the null hypothesis that the distributions have equal means with two-tailed \textit{p} value $< 0.0002$. For \rev{population 3}{training population 3 evaluated with population 3}, we consider it fails to reject the hypothesis ($p > 0.2$ for Adaptive with Generalist and $p > 0.6$ for Oracle and Generalist). However, this seems to be due not to the agent's failure to adapt but due to the Generalist already being almost optimal, within 0.1 point of the Oracle. The Adaptive score was, in fact, much closer to that of the Oracle than the Generalist. Note that for populations 1 and 2 there is more room for improvement between the Generalist and Oracle (roughly 0.5 and 0.9 point respectively).

The remaining scenarios are ``out-of-distribution'' scenarios, where the meta-agent interprets the observed behavioral features as if they were coming from one of the partners in its training population and adapts accordingly, but these features are coming from evaluation partners from a distinct, but similar, population. The out-of-distribution scenarios, therefore, are meant to test the robustness of the Adaptation meta-strategy to different evaluation partners.

The out-of-distribution scenarios had inconclusive, but slightly negative, results. In two scenarios, (1,3) and (2,1), the Generalist beat the Adaptive version by around 0.2 point. These scenarios had p-values around 0.001. In the remaining four scenarios, differences were deemed unlikely to be significant ($p>0.1$). This suggests that the Adaptive meta-strategy does not transfer particularly well out of population, but it also does not degrade performance by much.

We also see that, in out-of-distribution scenarios, the Oracle's advantage over the Adaptive and Generalist agents essentially disappears. It appears that knowing the correct niche of a partner does not convey much advantage for an Oracle with the wrong match-up table. 

    Finally, a closer look at \rev{column 2 of}{the scenarios with evaluation population 2 on} the table shows an unexpected result: both the Adaptive and Generalist instances trained with populations 1 and 3 beat in score all but the Oracle from population 2. While training with the correct population should be an advantage, recall that we use the training population as response pool as well. If population 2 has weaker agents in general, the agent using it as a response pool might be at an disadvantage. Further experiments, where the training and strategy pools are chosen independently, may shed further light on this finding.

\section{Discussion and Future Work}

Most game-playing agents in the literature, both for Hanabi and other games, are deployed with a ``frozen'' policy that displays little adaptation to other players, especially in-between games. This works well enough in competitive games or in games where it is possible to do well without modeling other players, Hanabi is a domain where an agent that is able to adapt its own behavior over short interactions with players would be desirable, especially for playing with humans. This ability could also be valuable outside the domain of game-playing agents, such as in mixed-initiative design systems, virtual assistants, etc.

In this paper, we presented a Hanabi ``Bayesian meta-agent'' that adapts to unknown partners within a small number of games. We use MAP-Elites to generate training populations that form a ``pool of hypotheses'' with high behavioral diversity. The meta-agent then collects, behavioral features from training match-ups between the training population and a pool of candidate response strategies. 

We evaluate the meta-agent by having it play short series of games with unknown partners. We use the collected behavioral features to perform Bayesian updates on a belief distribution that represents the belief that the unknown ad-hoc partner is using the same strategy as each of the agents in the pool of hypotheses. Finally, the meta-agent selects actions following the policy from its response pool that is thought to maximize score with partners sampled according to belief distribution.

\rev{The agent succeeds in our main objective: when being evaluated with the same population it was trained with, the Adaptive version improves its score from an initial ``Generalist'' meta-strategy in two of three scenarios. The third scenario is less clear, as the Generalist is already very close in performance to an optimal ``Oracle'' meta-strategy. It achieves this while keeping track of only three features during training and evaluation: the average Communicativeness, IPP and scores of each match-up.}{
In two of the three in-distribution scenarios, the Adaptive version slightly improves its score from an initial ``Generalist'' meta-strategy. The third scenario is less clear, as the Generalist is already very close in performance to an optimal ``Oracle'' meta-strategy. It achieves this while keeping track of only three features during training and evaluation: the average Communicativeness, IPP, and scores of each match-up.
}

The fact that self-play scores and pairwise scores are strongly correlated (see table~\ref{table:runs}) may affect our results, since, for the most  part, there's a single scale of performance for all agents rather than each agent's value being contextually dependent on each match-up. We know from~\cite{bard2020hanabi} that strong self-play agents can be bad at playing with others, but our populations largely don't reflect this.

For in-distribution performance, there are a few immediate avenues of improvement: first, we could improve  $R$ by training separate agents that play well with subsets of $H$ rather than simply using $R=H$. This would likely improve the score of all three versions (Oracle, Adaptive and Generalist) of the meta-agent. Second, we could drop the assumption that behavioral features in the match-ups between $R$ and $H$ are independent and normally distributed, and compute a better approximation of the real distribution from the training data. This would likely further bridge the gap between the Adaptive and Oracle versions. Third, we could calculate action probabilities directly from the policies of agents in $H$, rather than measuring behavior indirectly through behavioral features. 

 There might, however be a trade-off between improving same-distribution performance, where overfitting to the training data is desirable, and out-of-distribution performances, where overfitting is likely harmful. 

 The out-of-distribution results range from inconclusive to slightly negative. It is possible that we need behavioral features that better capture relevant strategy aspects. For example, neither of the current behavior dimensions captures whether the partner has a bias towards playing or discarding their oldest or newest card preferentially, which is common between humans. Adaptation intervals could also be defined contextually rather than by a fixed-duration: for example, the agent could attempt to adapt whenever it or its partner made a mistake.

\rev{}{Another open question is what happens when two agents simultaneously try to adapt to each other. In some exploratory experiments where an Adaptive meta-agent played with a copy of itself, both agents converged to a response neighboring the Generalist response. This might be because, in our populations, the Generalist is a near-optimal response to most agents in the pool. Different approaches to designing adaptive agents, however, could lead to interesting oscillatory behavior, since, at each step, the agents would be simultaneously adapting to each other's past behavior.}

Finally, we could use a larger or more varied training and evaluation pools, consisting of multiple runs of MAP-Elites, various hand-crafted agents, RL agents etc. An agent that could  adapt to such a diverse selection of agent would be an important next step towards cooperation with humans.

\section*{ACKNOWLEDGMENT}

Rodrigo Canaan, Andy Nealen and Julian Togelius gratefully acknowledge the financial support from Honda Research Institute Europe (HRI-EU).

\bibliographystyle{IEEEtran}
\bibliography{bibfile}

\begin{thebibliography}{10}
\providecommand{\url}[1]{#1}
\csname url@samestyle\endcsname
\providecommand{\newblock}{\relax}
\providecommand{\bibinfo}[2]{#2}
\providecommand{\BIBentrySTDinterwordspacing}{\spaceskip=0pt\relax}
\providecommand{\BIBentryALTinterwordstretchfactor}{4}
\providecommand{\BIBentryALTinterwordspacing}{\spaceskip=\fontdimen2\font plus
\BIBentryALTinterwordstretchfactor\fontdimen3\font minus
  \fontdimen4\font\relax}
\providecommand{\BIBforeignlanguage}[2]{{%
\expandafter\ifx\csname l@#1\endcsname\relax
\typeout{** WARNING: IEEEtran.bst: No hyphenation pattern has been}%
\typeout{** loaded for the language `#1'. Using the pattern for}%
\typeout{** the default language instead.}%
\else
\language=\csname l@#1\endcsname
\fi
#2}}
\providecommand{\BIBdecl}{\relax}
\BIBdecl

\bibitem{campbell2002deep}
M.~Campbell, A.~J. Hoane~Jr, and F.-h. Hsu, ``Deep blue,'' \emph{Artificial
  intelligence}, vol. 134, no. 1-2, pp. 57--83, 2002.

\bibitem{silver2016mastering}
D.~Silver, A.~Huang, C.~J. Maddison, A.~Guez, L.~Sifre, G.~Van Den~Driessche,
  J.~Schrittwieser, I.~Antonoglou, V.~Panneershelvam, M.~Lanctot \emph{et~al.},
  ``Mastering the game of {G}o with deep neural networks and tree search,''
  \emph{nature}, vol. 529, no. 7587, pp. 484--489, 2016.

\bibitem{vinyals2019grandmaster}
O.~Vinyals, I.~Babuschkin, W.~M. Czarnecki, M.~Mathieu, A.~Dudzik, J.~Chung,
  D.~H. Choi, R.~Powell, T.~Ewalds, P.~Georgiev \emph{et~al.}, ``Grandmaster
  level in {S}tar{C}raft {II} using multi-agent reinforcement learning,''
  \emph{Nature}, vol. 575, no. 7782, pp. 350--354, 2019.

\bibitem{canaan2019diverse}
R.~Canaan, J.~Togelius, A.~Nealen, and S.~Menzel, ``Diverse agents for ad-hoc
  cooperation in {H}anabi,'' in \emph{2019 IEEE Conference on Games
  (CoG)}.\hskip 1em plus 0.5em minus 0.4em\relax IEEE, 2019, pp. 1--8.

\bibitem{spielbgg}
\BIBentryALTinterwordspacing
BoardGameGeek, ``Spiel des jahres,'' access: 02/21/2021. [Online]. Available:
  \url{https://boardgamegeek.com/wiki/page/Spiel_des_Jahres}
\BIBentrySTDinterwordspacing

\bibitem{walton2017evaluating}
J.~Walton-Rivers, P.~R. Williams, R.~Bartle, D.~Perez-Liebana, and S.~M. Lucas,
  ``Evaluating and modelling {H}anabi-playing agents,'' in \emph{Evolutionary
  Computation (CEC), 2017 IEEE Congress on}.\hskip 1em plus 0.5em minus
  0.4em\relax IEEE, 2017, pp. 1382--1389.

\bibitem{bard2020hanabi}
N.~Bard, J.~N. Foerster, S.~Chandar, N.~Burch, M.~Lanctot, H.~F. Song,
  E.~Parisotto, V.~Dumoulin, S.~Moitra, E.~Hughes \emph{et~al.}, ``The {H}anabi
  challenge: A new frontier for {AI} research,'' \emph{Artificial
  Intelligence}, vol. 280, p. 103216, 2020.

\bibitem{hanabirules}
\BIBentryALTinterwordspacing
{R\&R Games}, ``Hanabi rules,'' access: 07/16/2021. [Online]. Available:
  \url{https://rnrgames.com/Content/RRGames/images/ProductRules/hanabiRules.PDF}
\BIBentrySTDinterwordspacing

\bibitem{cox2015make}
C.~Cox, J.~De~Silva, P.~Deorsey, F.~H. Kenter, T.~Retter, and J.~Tobin, ``How
  to make the perfect fireworks display: Two strategies for {H}anabi,''
  \emph{Mathematics Magazine}, vol.~88, no.~5, pp. 323--336, 2015.

\bibitem{bouzy2017playing}
B.~Bouzy, ``Playing {H}anabi near-optimally,'' in \emph{Advances in Computer
  Games}.\hskip 1em plus 0.5em minus 0.4em\relax Springer, 2017, pp. 51--62.

\bibitem{foerster2019bayesian}
J.~Foerster, F.~Song, E.~Hughes, N.~Burch, I.~Dunning, S.~Whiteson,
  M.~Botvinick, and M.~Bowling, ``Bayesian action decoder for deep multi-agent
  reinforcement learning,'' in \emph{International Conference on Machine
  Learning}.\hskip 1em plus 0.5em minus 0.4em\relax PMLR, 2019, pp. 1942--1951.

\bibitem{hu2019simplified}
H.~Hu and J.~N. Foerster, ``Simplified action decoder for deep multi-agent
  reinforcement learning,'' in \emph{International Conference on Learning
  Representations}, 2019.

\bibitem{osawa2015solving}
H.~Osawa, ``Solving {H}anabi: Estimating hands by opponent's actions in
  cooperative game with incomplete information.'' in \emph{AAAI workshop:
  Computer Poker and Imperfect Information}, 2015, pp. 37--43.

\bibitem{van2016aspects}
M.~J. van~den Bergh, A.~Hommelberg, W.~A. Kosters, and F.~M. Spieksma,
  ``Aspects of the cooperative card game {H}anabi,'' in \emph{Benelux
  Conference on Artificial Intelligence}.\hskip 1em plus 0.5em minus
  0.4em\relax Springer, 2016, pp. 93--105.

\bibitem{canaan2020behavioral}
R.~Canaan, X.~Gao, Y.~Chung, J.~Togelius, A.~Nealen, and S.~Menzel,
  ``Behavioral evaluation of {H}anabi {R}ainbow {DQN} agents and rule-based
  agents,'' in \emph{Proceedings of the AAAI Conference on Artificial
  Intelligence and Interactive Digital Entertainment}, vol.~16, no.~1, 2020,
  pp. 31--37.

\bibitem{walton20192018}
J.~Walton-Rivers, P.~R. Williams, and R.~Bartle, ``The 2018 {H}anabi
  competition,'' in \emph{2019 IEEE Conference on Games (CoG)}.\hskip 1em plus
  0.5em minus 0.4em\relax IEEE, 2019, pp. 1--8.

\bibitem{canaan2018evolving}
R.~Canaan, H.~Shen, R.~Torrado, J.~Togelius, A.~Nealen, and S.~Menzel,
  ``Evolving agents for the {H}anabi 2018 {CIG} competition,'' in \emph{2018
  IEEE Conference on Computational Intelligence and Games (CIG)}.\hskip 1em
  plus 0.5em minus 0.4em\relax IEEE, 2018, pp. 1--8.

\bibitem{eger2017intentional}
M.~Eger, C.~Martens, and M.~Alfaro~C{\'o}rdoba, ``An intentional {AI} for
  {H}anabi,'' in \emph{Computational Intelligence and Games (CIG), 2017 IEEE
  Conference on}.\hskip 1em plus 0.5em minus 0.4em\relax IEEE, 2017, pp.
  68--75.

\bibitem{liang2019implicit}
C.~Liang, J.~Proft, E.~Andersen, and R.~A. Knepper, ``Implicit communication of
  actionable information in human-{AI} teams,'' in \emph{Proceedings of the
  2019 CHI Conference on Human Factors in Computing Systems}, 2019, pp. 1--13.

\bibitem{grice1975logic}
H.~P. Grice, ``Logic and conversation,'' in \emph{Speech acts}.\hskip 1em plus
  0.5em minus 0.4em\relax Brill, 1975, pp. 41--58.

\bibitem{lerer2020improving}
A.~Lerer, H.~Hu, J.~Foerster, and N.~Brown, ``Improving policies via search in
  cooperative partially observable games,'' in \emph{Proceedings of the AAAI
  Conference on Artificial Intelligence}, vol.~34, no.~05, 2020, pp.
  7187--7194.

\bibitem{hu2020other}
H.~Hu, A.~Lerer, A.~Peysakhovich, and J.~Foerster, ``“other-play” for
  zero-shot coordination,'' in \emph{International Conference on Machine
  Learning}.\hskip 1em plus 0.5em minus 0.4em\relax PMLR, 2020, pp. 4399--4410.

\bibitem{goodman2019re}
J.~Goodman, ``Re-determinizing {MCTS} in {H}anabi,'' in \emph{2019 IEEE
  Conference on Games (CoG)}.\hskip 1em plus 0.5em minus 0.4em\relax IEEE,
  2019, pp. 1--8.

\bibitem{stone2010ad}
P.~Stone, G.~A. Kaminka, S.~Kraus, and J.~S. Rosenschein, ``Ad hoc autonomous
  agent teams: Collaboration without pre-coordination,'' in \emph{Twenty-Fourth
  AAAI Conference on Artificial Intelligence}, 2010.

\bibitem{pugh2016quality}
J.~K. Pugh, L.~B. Soros, and K.~O. Stanley, ``Quality diversity: A new frontier
  for evolutionary computation,'' \emph{Frontiers in Robotics and AI}, vol.~3,
  p.~40, 2016.

\bibitem{lehman2011abandoning}
J.~Lehman and K.~O. Stanley, ``Abandoning objectives: Evolution through the
  search for novelty alone,'' \emph{Evolutionary computation}, vol.~19, no.~2,
  pp. 189--223, 2011.

\bibitem{deb2014multi}
K.~Deb, ``Multi-objective optimization,'' in \emph{Search methodologies}.\hskip
  1em plus 0.5em minus 0.4em\relax Springer, 2014, pp. 403--449.

\bibitem{mouret2015illuminating}
J.-B. Mouret and J.~Clune, ``Illuminating search spaces by mapping elites,''
  \emph{arXiv preprint arXiv:1504.04909}, 2015.

\bibitem{cully2015robots}
A.~Cully, J.~Clune, D.~Tarapore, and J.-B. Mouret, ``Robots that can adapt like
  animals,'' \emph{Nature}, vol. 521, no. 7553, p. 503, 2015.

\bibitem{fontaine2019mapping}
M.~C. Fontaine, S.~Lee, L.~B. Soros, F.~de~Mesentier~Silva, J.~Togelius, and
  A.~K. Hoover, ``Mapping {H}earthstone deck spaces through map-elites with
  sliding boundaries,'' in \emph{Proceedings of The Genetic and Evolutionary
  Computation Conference}, 2019, pp. 161--169.

\bibitem{fontaine2020covariance}
M.~C. Fontaine, J.~Togelius, S.~Nikolaidis, and A.~K. Hoover, ``Covariance
  matrix adaptation for the rapid illumination of behavior space,'' in
  \emph{Proceedings of the 2020 genetic and evolutionary computation
  conference}, 2020, pp. 94--102.

\bibitem{hansen2016cma}
N.~Hansen, ``The {CMA} evolution strategy: A tutorial,'' \emph{arXiv preprint
  arXiv:1604.00772}, 2016.

\bibitem{tesauro2003extending}
G.~Tesauro, ``Extending {Q}-learning to general adaptive multi-agent systems,''
  in \emph{Advances in neural information processing systems}.\hskip 1em plus
  0.5em minus 0.4em\relax Citeseer, 2003, p. None.

\bibitem{chalkiadakis2003coordination}
G.~Chalkiadakis and C.~Boutilier, ``Coordination in multiagent reinforcement
  learning: A bayesian approach,'' in \emph{Proceedings of the second
  international joint conference on Autonomous agents and multiagent systems},
  2003, pp. 709--716.

\bibitem{hoang2013general}
T.~N. Hoang and K.~H. Low, ``A general framework for interacting
  bayes-optimally with self-interested agents using arbitrary parametric model
  and model prior,'' in \emph{Twenty-Third International Joint Conference on
  Artificial Intelligence}, 2013.

\bibitem{wang2020too}
R.~E. Wang, S.~A. Wu, J.~A. Evans, J.~B. Tenenbaum, D.~C. Parkes, and
  M.~Kleiman-Weiner, ``Too many cooks: Coordinating multi-agent collaboration
  through inverse planning,'' in \emph{Proceedings of the 19th International
  Conference on Autonomous Agents and MultiAgent Systems}, 2020, pp.
  2032--2034.

\bibitem{zintgraf2021deep}
L.~Zintgraf, S.~Devlin, K.~Ciosek, S.~Whiteson, and K.~Hofmann, ``Deep
  interactive bayesian reinforcement learning via meta-learning,'' in
  \emph{Proceedings of the 20th International Conference on Autonomous Agents
  and MultiAgent Systems}, 2021, pp. 1712--1714.

\bibitem{rish2001empirical}
I.~Rish \emph{et~al.}, ``An empirical study of the naive bayes classifier,'' in
  \emph{IJCAI 2001 workshop on empirical methods in artificial intelligence},
  vol.~3, no.~22, 2001, pp. 41--46.

\bibitem{welch1947generalization}
B.~L. Welch, ``The generalization of student's' problem when several different
  population variances are involved,'' \emph{Biometrika}, vol.~34, no. 1/2, pp.
  28--35, 1947.

\end{thebibliography}

\end{document}